\documentclass{ws-jmrr}
\usepackage[sort,compress,super]{cite}

\usepackage{tabularx}
\usepackage{multirow}
\usepackage{booktabs}

\usepackage{xcolor, soul}

\begin{document}

\catchline{0}{0}{2013}{}{}

\markboth{Ziheng Wang, Conor Perreault, Xi Liu, Anthony Jarc}{Automatic Detection of Out-of-body Frames in Surgical Videos for Privacy Protection Using Self-supervised Learning and Minimal Labels}
\title{Automatic Detection of Out-of-body Frames in Surgical Videos for Privacy Protection Using Self-supervised Learning and Minimal Labels}

\author{Ziheng Wang$^{a,}$\footnote{5655 Spalding Drive, Norcross, GA 30091\\
E-mail: Ziheng.Wang@intusurg.com}\ , Xi Liu$^b$, Conor Perreault$^b$, Anthony Jarc$^b$}

\maketitle

\begin{abstract} 
Endoscopic video recordings are widely used in minimally invasive robot-assisted surgery, but when the endoscope is outside the patient's body, it can capture irrelevant segments that may contain sensitive information. To address this, we propose a framework that accurately detects out-of-body frames in surgical videos by leveraging self-supervision with minimal data labels. We use a massive amount of unlabeled endoscopic images to learn meaningful representations in a self-supervised manner.
Our approach, which involves pre-training on an auxiliary task and fine-tuning with limited supervision, outperforms previous methods for detecting out-of-body frames in surgical videos captured from da Vinci X and Xi surgical systems. The average F1 scores range from $96.00\pm2.09$ to $98.02\pm1.00$. Remarkably, using only $5\%$ of the training labels, our approach still maintains an average F1 score performance above $97$, outperforming fully-supervised methods with $95\%$ fewer labels.
These results demonstrate the potential of our framework to facilitate the safe handling of surgical video recordings and enhance data privacy protection in minimally invasive surgery.


\end{abstract}

\section{Introduction}
\begin{figure*}[tb]
\centering
\includegraphics[width= 0.95\linewidth,clip,trim=0pt 0pt 0pt 0pt]{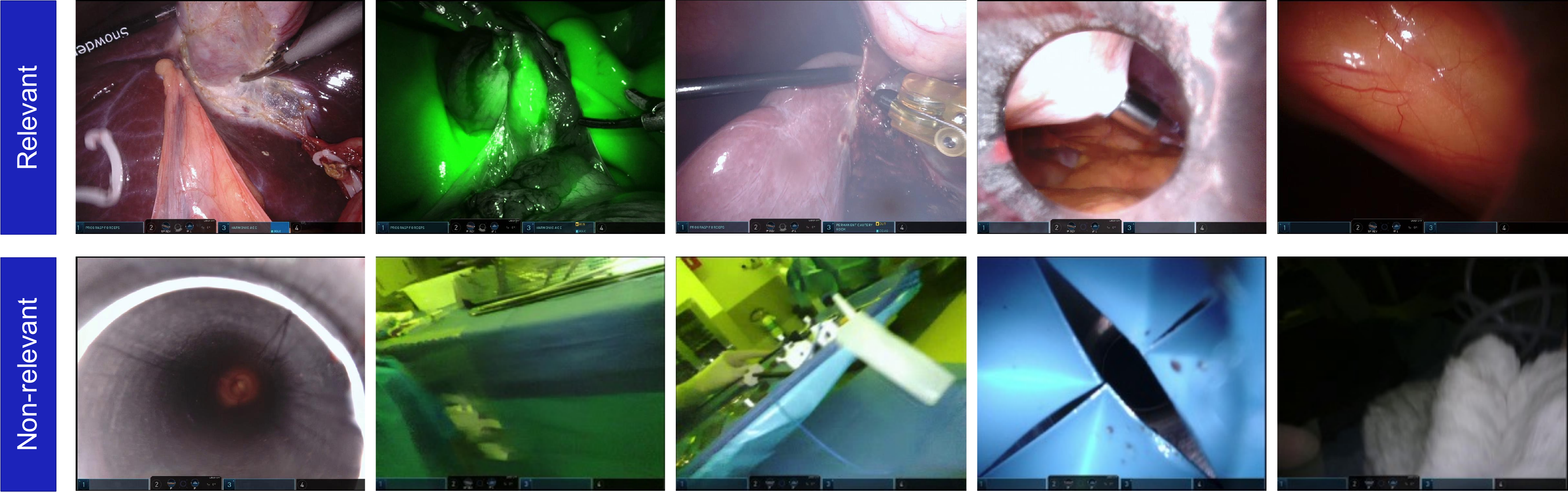}
\caption{Example of surgically relevant and irrelevant frames in video recordings of robot-assisted surgery. Several visual challenges present in frames of both relevant and irrelevant segments, such as the presence of smoke and motion blur. Clinically relevant frames also have a wide range of color/light variances between regular white-light and fluorescent imaging. Clinically irrelevant frames potentially contain privacy-sensitive information of patients, clinicians, instruments, etc. in the operating room.}
\label{fig_PHI_examples}
\end{figure*}

Surgical video recordings are an essential asset for surgeons and medical institutions to improve surgical education, safety, and liability~\cite{maier2017surgical,vedula2017surgical,maier2018surgical}. Although valuable for enabling new technologies and insights~\cite{rivas2021review,maier2018surgical}, the surgical data might unintentionally capture identifiable images of the patient or operating room (OR) personnel, health records, and other sensitive identifiers when the endoscope is removed from the patient’s body during a procedure. It is imperative to identify and remove any clinically irrelevant and potentially sensitive out-of-body segments so that videos are stored and used responsibly.

One commonly used approach to protect patient data privacy involves manual video editing, where annotators or clinicians review surgical recordings frame by frame to remove sensitive segments while preserving critical aspects of the surgery. However, this process is time-consuming and prone to human error, and there is a risk of missing short segments of sensitive frames that occur randomly during a surgical procedure. Additionally, it is not feasible to use manual editing for large amounts of full-length video recordings from surgeries that can last for hours. Furthermore, sensitive segments can be easily overlooked when the endoscope is quickly removed for camera cleaning or an instrument exchange, making it challenging to ensure complete data privacy protection.

With the recent development of surgical data science, computer-assisted methods have shown potential to automatically detect irrelevant frames in surgical videos. 
Initial approaches use machine learning algorithms based on low-level hand-crafted visual features of endoscopic images. These methods depend on extracting highly engineered image representations, such as color histograms, image texture, contour, and pixel brightness, etc.~\cite{oh2007informative,akgul2011content,stanek2012automatic,munzer2013relevance,atasoy2011endoscopic}. Stanek et al. proposed to use simple red channel measures in image RGB color space to detect out-of-body frames and further used them to identify the start and end of endoscopic procedures~\cite{stanek2012automatic}. M{\"u}nzer et al. considered using a set of visual representations as indicators to identify irrelevant frames, including hue distribution in HSV color space, Difference of Gaussians (DoG), and averaged pixel intensity~\cite{munzer2013relevance}. To detect irrelevant frames in gastro-intestinal endoscopic videos, Atasoy et al. proposed to calculate representation vectors based on the image power spectrum and built an unsupervised model using K-means clustering on extracted representations~\cite{atasoy2011endoscopic}. More recently, artificial intelligence systems using deep neural networks have been broadly pursued for image classification tasks~\cite{lecun2015deep,goodfellow2016deep}. One common approach is to build a fully supervised network from labeled datasets and learn domain-specific visual features of surgical images based on convolutional neural networks (CNNs). 
Given an adequate amount of annotations for model training, the features learned in a fully-supervised manner are able to yield high accuracy and have shown great success in solving problems for surgical skill assessment~\cite{funke2019video,wang2018deep}, surgical activity recognition~\cite{yoshida2022spatiotemporal,golany2022artificial}, segmentation~\cite{huang2022surgical, allan20192017}, tool localization~\cite{liu2020anchor,mei2019detection}, etc. 
However, building such supervised models is challenging as it requires a large amount of data and human annotations. In reality, the availability of annotations is often limited due to the amount of time it takes for an expert to review and annotate the individual segments of a procedure.
To overcome this, several works have been proposed to leverage semi-supervised learning that require data with either noisy or fewer labels~\cite{chapelle2009semi}. 
Zohar et al. used a semi-supervised method to learn representations from noisy labels~\cite{zohar2020accurate}. Instead of using accurate labels of surgically relevant/irrelevant images, they proposed to only annotate the start and end timestamps of a procedure and treated all images during the procedure as surgically relevant and all other frames as irrelevant. A deep neural network was then trained iteratively to learn from these noisy labels. While their approach potentially reduces the amount of human annotations required, annotating the start and end of a procedure is not an easy process and still requires expert knowledge as well as reviewing the full procedure. Also, variations in noisy labels and an iterative training process can negatively impact model performance. 
Another approach is to remove identifiable patient or surgeon faces from surgical videos based on face detection. A semi-supervised learning method was proposed to iteratively train a neural network on non-annotated clinical data~\cite{issenhuth2019face,flouty2018faceoff}. Though promising, this approach only focuses on face areas, and other sensitive entities that are typically present in surgical recordings, such as instruments, text, and badges, are ignored. 
Therefore, it is highly desirable to explore more effective approaches that can build high-performing models given limited label availability, reducing the need for human annotations. 

One area of research in the general image processing field focuses on self-supervised learning (SSL), which aims to to learn rich visual features in a domain-agnostic manner~\cite{jing2020self,kolesnikov2019revisiting,caron2018deep}. Rather than using domain-specific labels for supervision, self-supervised learning utilizes unlabeled data and generates a supervisory signal for modeling. This process is done by choosing a pseudo-supervised auxiliary task that applies a transformation on images and requires the model to predict the pseudo labels of the corresponding transformed image~\cite{chen2020adversarial}. Several simple yet effective transformations have been proposed to exploit different image properties by training models in auxiliary tasks, such as rotations~\cite{gidaris2018unsupervised}, color permutation~\cite{lee2020self}, colorization~\cite{deshpande2015learning,goyal2019scaling}, jigsaw puzzle~\cite{goyal2019scaling,misra2020self}, affine warpings~\cite{novotny2018self}, and inpainting~\cite{pathak2016context}. As a result, the representations learned by solving these auxiliary tasks can then be adapted to downstream classification tasks by fine-tuning the parameters of an existing architecture. However, the success of self-supervision in detecting irrelevant frames in endoscopic videos, a domain-specific problem, remains unclear given the significant differences between surgical endoscopic data and generic images, and it is not well understood how a model would perform when only limited data labels are available for this task.

\begin{figure*}[tb]
\centering
\includegraphics[width= 0.95\linewidth,clip ,trim=0pt 0pt 0pt 0pt]{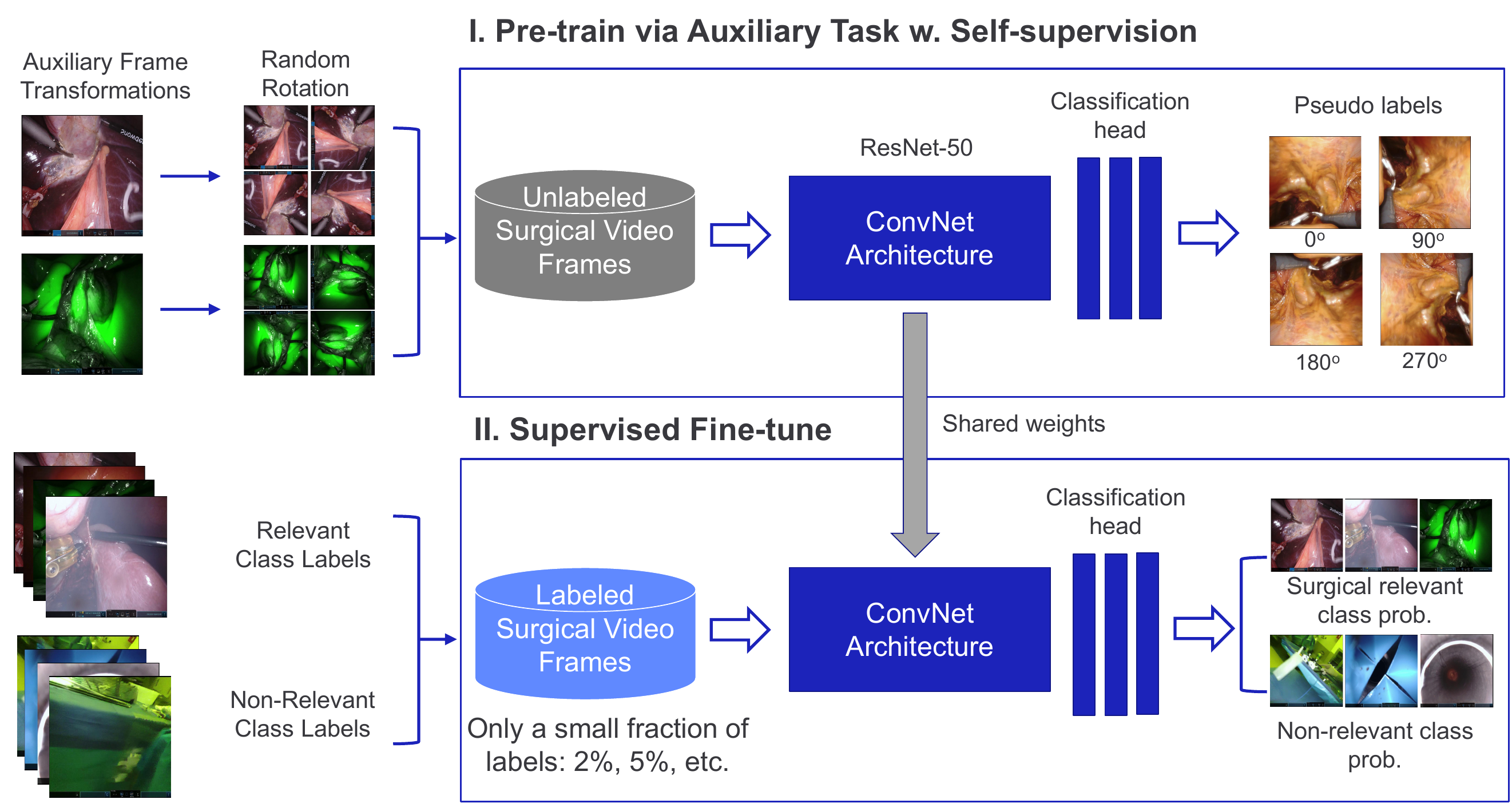}
\caption{A framework utilizing self-supervised learning (SSL) and minimal labels to accurately detect clinically irrelevant segments in surgical videos. The framework consists of two main stages: (1) pre-training on large-scale endoscopic images via SSL without the use of any labels. Specifically, we adopted a rotation-based SSL method where the model learns latent representations through an auxiliary task of recognizing random rotation transformations of an input image ($0^\circ$, $90^\circ$, $180^\circ$, and $270^\circ$), and (2) fine-tuning with limited supervision, where A model is trained to predict surgically relevant/irrelevant classes using only a small set of labeled samples, as annotations are typically challenging to generate.}
\label{fig_workflow}
\end{figure*}

In this study, we implement and evaluate a framework that utilizes self-supervised learning to accurately detect irrelevant segments in endoscopic surgical videos. Throughout a series of experiments on the clinical videos collected from a variety of robot-assisted surgeries, we have shown that it is feasible to build privacy-preserving models with very limited annotations.  
Unlike fully supervised approaches, the framework learns meaningful representations through self-supervision, without initially requiring any data labels. We believe this work addresses a critical need for privacy-preserving AI in the clinical community, and can improve the safe and efficient handling of surgical video recordings for a growing number of surgical data science applications.
Our main contributions are:
\begin{itemize}
\item Implementation and evaluation of a framework that leverages self-supervised learning with minimal labels to accurately detect out-of-body frames in surgical videos;
\item Comparison and benchmarking of out-of-body frame detection performance with varying availability of data labels;
\item Demonstrating that our resulting models can match and even outperform fully-supervised models with $95\%$ fewer labels, indicating the potential of our framework to reduce the need for large amounts of data labels and facilitate practical deployment in real-world clinical settings.
\end{itemize}

\section{Methodology}
The goal of our work is to accurately classify irrelevant segments throughout the full-length surgical video. We formulate the problem as a binary classification task where the out-of-body, surgically irrelevant video frame containing sensitive information is considered as the positive class and the in-body, surgically relevant video frame is considered the negative class. The final output is the scrubbed surgical video wherein any video segment containing irrelevant frames is removed. 

Fig.~\ref{fig_workflow} shows the overview of the framework design. 
The framework consists of two main stages: (1) self-supervised pre-training via an auxiliary task without labels and (2) supervised fine-tuning with a set of labeled samples. In this work, we adopt a prominent self-supervised learning method of recognizing rotation transformations to learn image features~\cite{jing2020self}. 
The rotation-based SSL method is used in this work because intra-operative surgical videos are generally recorded from distinct egocentric views. More specifically, the endoscope is usually positioned at a non-fixed angle and the viewpoint of the camera can be changed fairly randomly during a procedure. We hypothesize that applying rotation-like transformations to the input images for pre-training simulates the stochastic nature of endoscope motion and ultimately encourages the model to learn meaningful endoscope-specific representations. 

\subsection{Auxiliary Pre-training without Labels}

Learning informative visual features from images is crucial for achieving strong performance on downstream classification tasks. In contrast to supervised learning which relies on actual labels, we employ a self-supervised learning approach using rotation transformations to generate the supervisory signal.
To implement this approach, we follow a similar setup as in~\cite{gidaris2018unsupervised}, randomly rotating images from the entire unlabeled data and training a model to predict the corresponding rotation angle. Specifically, we consider four rotation angels for each input image: $0^\circ$, $90^\circ$, $180^\circ$, and $270^\circ$, chosen to maximize  visual differences and promote effective pre-training. We defined the loss as follows:
 \begin{equation}
    \mathcal{L}_{rot} = \frac{1}{R} \mathop{\sum_{r\in R}} \sum_{x\in D_{u}} \mathcal{L} (f_\theta(x_r), r)
 \end{equation}
where $R$ is the set of four rotation degrees $\{0^\circ, 90^\circ, 180^\circ, 270^\circ\}$, $x$ is an image sampled from the unlabeled training set $D_{u}$, $x_r$ is the image $x$ rotated by a degree $r\in R$, $f_\theta(\cdot)$ is the model with parameters $\theta$, and $\mathcal{L}_{rot}$ is the cross entropy loss of rotation classification.
On top of the feature representations, a classification head that maps these vectors into the classification space is added in pre-training task. We use a multi-layer perceptron (MLP) with two hidden layers (MLP-2) to obtain the softmax probability distribution of the predicted rotation angle. A total of 512-unit neurons are used in the first hidden layer. A ReLU non-linearity and random dropout with rate of 0.1 is applied between the two hidden layers.

\subsection{Supervised Fine-tuning with Limited Supervision}
Once the pre-trained model is obtained, we apply transfer learning and fine-tune the architecture on top of the representations learned by the pre-trained backbone on a set of supervised labels. A new classification head with the two-hidden-layer MLP-2 is added to the network to replace the classification head used during the pre-training. The aim here is to build a binary classifier that can determine whether an endoscopic video frame is inside or outside of the body during surgery. The loss in fine-tuning is defined as:
 \begin{equation}
    \mathcal{L} =  \frac{1}{Y}\mathop{\sum_{y\in Y}} \sum_{x\in D_{l}} \mathcal{L} (f_\theta^\prime(x), y)
 \end{equation}
where $x$ is an image in the labeled dataset $D_{l}$ and y $\in Y$ is the corresponding class label of interest $\{relevant, irrelevant\}$, $f_\theta^\prime(\cdot)$ is the fine-tuned model with adjusted parameters $\theta$, and $\mathcal{L}$ is the standard cross-entropy loss.  
For experiments, we construct the labeled datset $D_{l}$ by randomly sampling $2\%$, $5\%$, $10\%$, and $15\%$ of all the labeled images in training set, and fine-tune the whole network on the available labeled data.
Details of the model training and validation are presented in Section~\ref{sec_implementation}.

\subsection{Dataset}
Our dataset was collected from multiple procedures in robot-assisted surgery using either da Vinci X or da Vinci Xi surgical systems (Intuitive Surgical Inc., Sunnyvale, CA), collected from different medical facilities. It contained a total of 145 endoscopic videos from four types of robot-assisted surgical procedures, including cholecystectomy, lobectomy, hysterectomy, and hemicolectomy. Raw video was recorded at 30 frames per second (fps) with an image resolution of $1280\times720$ pixels. 
We downsampled each raw video to 1 fps and considered each video frame as an individual sample. Given the dataset of 145 surgical videos, a total of 78377 frames (72666 relevant, 5711 irrelevant) were extracted for the training, validation, and testing. 
The start and end time of each relevant and irrelevant segment were first manually annotated by five expert specialists using an in-house annotation tool. Then, frame-wise labels for every second were derived from the labels of each video segment. 

For our experiments, we randomly split the dataset into training and evaluation (validation and testing) sets based on different surgical cases, and a strict split is enforced across experiments. In order to obtain a robust estimation of model performance, we followed a five-fold cross validation setting and the same experiment was repeated five times. In each fold, a random subset of $45\%$ of the surgical cases were selected for training, the remaining $20\%$ and $35\%$ of the surgical cases were used for validation and testing, respectively.  
For consistency, the same split and random seed was applied for all experiments in each individual fold to obtain reproducible benchmarks when comparing models. We chose the best model weights based on the lowest cross-entropy loss on the validation set, and evaluated performance on the testing set in each fold. We reported the average and standard deviation of metrics on the testing set across five folds as the model performance measures.

Table~\ref{tab:datasets} shows the summary of the surgical video dataset and descriptive statistics of data used for training, validation and testing in every fold of experiments. 

\begin{table}
\tbl{Summary of the surgical video dataset containing recordings collected in robot-assisted surgery. For experiments, we followed a random five-fold evaluation setting and validated models against testing sets in the five different splits. In each fold, a random split of surgical cases was performed for training (45$\%$), validation (20$\%$), and testing (35$\%$). \label{tab:datasets}}
{\begin{tabular}{lllllllll}
\Hline
\parbox{2cm}{Cross-validation} & Category & Num. Videos & \parbox{1.6cm}{Num.\\Relevant\\ Frames } & \parbox{1.6cm}{Num. Irrelevant\\Frames} & \parbox{1.5cm}{Num. Relevant Segments} & \parbox{1.5cm}{Num. Irrelevant Segments} &\parbox{1.5cm}{Ratio\\(Irrelevant:Relevant)}\\ 
\hline
\multirow{1}{*}{Fold I} & Train & 61 & 31788 & 2726 & 92 & 44 & 0.0858\\
& Validation & 37 & 21396 & 1605 & 57 & 27 & 0.0750\\
& Test & 47 & 25193 & 1380 & 58 & 18 & 0.0548 \\
\midrule
\multirow{2}{*}{Fold II} & Train & 70 & 37357 & 1835 & 97 & 39 & 0.0491\\
& Validation & 33 & 19926 & 748 & 43 & 13 & 0.0375\\
& Test & 42 & 21094 & 3128 & 67 & 37 & 0.1483\\
\midrule
\multirow{3}{*}{Fold III} & Train & 75 & 38780 & 3691 & 110 & 49 & 0.0952\\
& Validation & 18 & 10495 & 1037 & 31 & 20 & 0.0988\\
& Test & 52 & 29102 & 983 & 66 & 20 & 0.0338\\
\midrule
\multirow{4}{*}{Fold IV} & Train & 65 & 32014 & 4048 & 100 & 53 & 0.1264\\
& Validation & 36  & 21417 & 751 & 46 & 14 & 0.0351\\
& Test & 44 & 24946 & 912 & 61 & 22 & 0.0366\\
\midrule
\multirow{5}{*}{Fold V} & Train & 76 & 40496 & 2670 & 107 & 44 & 0.0659\\
& Validation & 29 & 15240 & 1477 & 42 & 20 & 0.0969\\
& Test & 40 & 22641 & 1564 & 58 & 25 & 0.0691\\
\hline
Summary & Total & 145 & 78377 & 5711 & 207 & 89 & 0.0729\\
\Hline
\end{tabular}}
\end{table}

\subsection{Implementation Details}
\label{sec_implementation}
\paragraph{\textbf{Image Preprocessing}} 
We uniformly preprocessed the image frames for the input to the framework after downsampling the original video.
Each frame was first centrally cropped to a size of $640\times640$ pixels. For validation and testing data, we further resized the frame to $224\times224$. All images in the dataset were normalized using the mean (mean=$[0.485, 0.456, 0.406]$) and standard deviation (std=$[0.229, 0.224, 0.225]$) values, following common practice~\cite{he2016deep}.

\begin{figure}[tb]
\centering
\includegraphics[width=0.5\linewidth,clip ,trim=0pt 25pt 0pt 0pt]{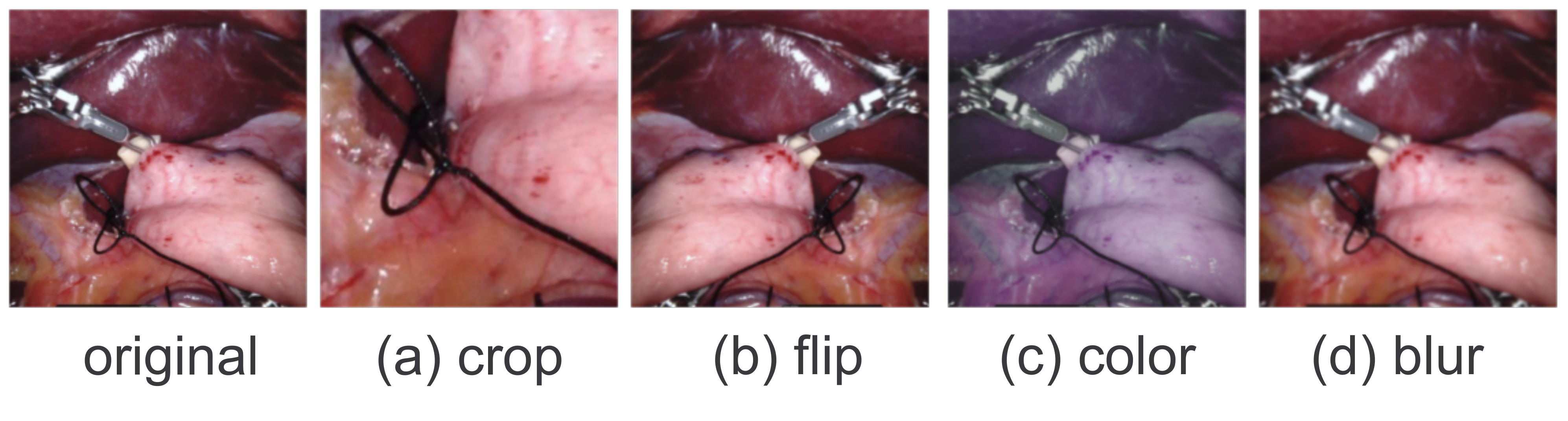}
\caption{An example of an image with different data augmentations applied. Details are described in Section \ref{sec:augment}.}
\label{fig_augment}
\end{figure}

\paragraph{\textbf{Data Augmentation}}
\label{sec:augment}
Stochastic data augmentation plays an important role in improving the quality of learned image representations and can largely affect the performance of self-supervised learning. Similar to the work~\cite{chen2020simple}, we adopted four sequential augmentations for training either self-supervised or fully-supervised models: random cropping, random flipping, random color distortion, and random Gaussian blur. As shown in Fig.~\ref{fig_augment}, we random cropped the input image ($640\times640$) with a uniform scale from $0.08$ to $1.00$ and a random aspect ratio from $3/4$ to $4/3$, and then resize to $224\times224$. Then we flipped images horizontally with $50\%$ probability. For color distortion, we applied a random change of image brightness, contrast, saturation with a scale from 0.8 to 1.2, and a random change of hue with a scale from -0.1 to 0.1. The color distortion was applied to the image $90\%$ of the time. Lastly, we applied Gaussian blur using a Gaussian kernel with the kernel size of 3 and a random $\sigma$ from 0.1 to 2.0. No data augmentation was applied for the validation and testing sets.

\paragraph{\textbf{Training}}
To pre-train our model, we utilized unlabeled samples and the widely-used ResNet-50 architecture as our backbone. This choice was based on its popularity and effectiveness in self-supervised frames, as well as its success in the field of surgical data science, where it has been adopted for surgical activity recognition~\cite{zia2018surgical}. We trained the architecture on all available images in the training set, without the use of any labels. Our training process was performed using four GeForce RTX 2080 GPUs, with a batch size of 80, for a total of 150 epochs. We used the Adam optimizer with a fixed learning rate of 0.001 throughout all epochs.

For fine-tuning, we implemented a learning rate warm-up strategy using a scheduler. Initially, the learning rate was set to 0.001, and then linearly decayed by $10\%$ after the first $25\%$, $50\%$, and $75\%$ epochs. The minimum learning rate of $1e^{-6}$ was reached at the end of training. We used a batch size of 100 and reduced the total number of epochs to 40. We used a batch size of 100 and reduced the total number of epochs to 40. 

To establish a baseline, we simply fine-tuned the model using all available labels in the training set and with a randomly sampled subset of labels (including $2\%$, $5\%$, $10\%$, and $15\%$ of the total training set). To ensure reproducibility and fair comparison between models, we used a random number generator with a predefined global seed across individual experiments.
It's worth noting that we aimed to alleviate class imbalance and obtain a more robust model by regularizing the multicategorical cross-entropy loss with empirical class weights of 0.15 and 0.85 for the negative and positive classes, respectively. 

To evaluate the performance of the rotation-based self-supervision algorithm, we also trained self-supervised models using two other standard self-supervision algorithms: Jigsaw~\cite{noroozi2016unsupervised} and SimCLR~\cite{chen2020simple}. To ensure a fair comparison of fine-tuning capabilities, we trained the self-supervised models using the same hyperparameters as were successful in the original papers to propose the algorithms. We completed the fine-tuning process with the same hyperparameters used in all of our fine-tuning and supervised experiments, allowing us to use the best hyperparameters for feature extraction while ensuring a fair comparison of the algorithms' fine-tuning performance.
We utilized the VISSL library's~\cite{goyal2021vissl} implementation of these algorithms.

For the evaluation of the entire self-supervised framework, we built fully-supervised models using a standard ResNet-50 architecture as the backbone with a 2-layer MLP classification head (MLP-2) with $N$ logit units, where $N$ corresponds to the number of output classes, $N=2$. 
The ResNet-50 backbone was initialized using either pre-trained ImageNet weights or random initialized weights (train from scratch). 
We created a fully-supervised baseline using $100\%$ of the training labels. Similarly, we reported performance when training fully-supervised models with limited labels ($2\%$, $5\%$, $10\%$, and $15\%$) for comparisons. The same set of image augmentations - namely random crop, random horizontal flip, color distortion, and Gaussian blur - was used. For fair comparison, we shared the same learning rate schedule, loss regularization, batch size, and epochs to train fully-supervised models as was used for fine-tuning self-supervised models.

\paragraph{\textbf{Evaluation}}
For evaluating the overall performance of the detection model, we measured the macro-averaged F1-score by calculating the harmonic mean of F1 scores with respect to each class~\cite{sokolova2009systematic}, as shown in Eq.~\ref{eq:mF1}.
As it is essential to accurately identify sensitive out-of-body images, we also specifically reported the standard accuracy metrics~\cite{sokolova2009systematic}, precision, recall, and F1 score (F1) of the positive (surgically irrelevant) class. 
\begin{equation}
mF1 = mean(F1_{relevant}, F1_{irrelevant})
\label{eq:mF1}
\end{equation}
Note that the class observations in surgical videos are most likely highly imbalanced, and only a small fraction of images are positives (\textit{i.e.}, surgically irrelevant), as shown in Table~\ref{tab:datasets}. Since our priority is to accurately detect the minority class, calculating the macro-averaged F1 results in a larger penalisation when a model does not perform well with the minority class, and thus provides a better evaluation of the model's capacity for our focus.

\begin{table}
\tbl{Numbers of labeled frames in a training set given varied label percentages in the five-fold experiments. The model trained with $100\%$ of labels is considered a baseline given full availability of training samples.
\label{tab:trainset_percentage_summary}}
{
\begin{tabular}{lccccc}
\Hline
\multirow{2}{*}{Cross-validation}  & \multicolumn{5}{c}{Label Percentage}  \\
&  2$\%$ & 5$\%$ & 10$\%$ & 15$\%$ & 100\%\\
\hline
Fold I & 635 & 1589 & 3178 & 4768  & 31788 \\
Fold II & 747 & 1867 & 3735 & 5603  & 37357 \\
Fold III & 849 & 1939 & 3878 & 5817  & 38780 \\
Fold IV & 640 & 1600 & 3201 & 4802  & 32014  \\
Fold V & 809 & 2024 & 4049 & 6074  & 40496 \\
\hline
Average & 736 & 1803 & 3608 & 5412  & 36087 \\
\Hline
\end{tabular}}
\end{table}

\begin{figure}[tb]
\centering
\includegraphics[width= 0.45\linewidth,clip,trim=0pt 0pt 0pt 0pt]{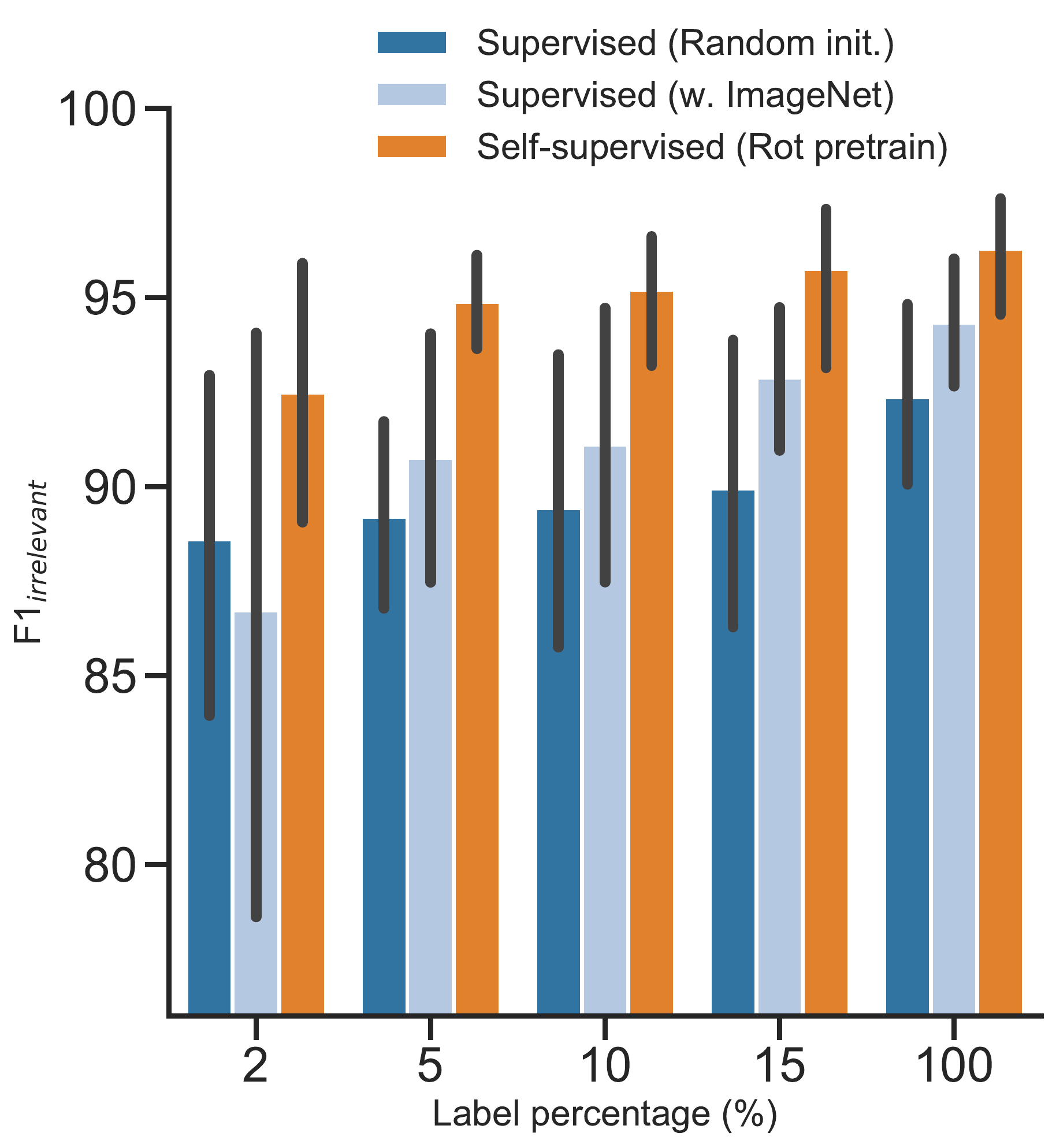}
\caption{F1 accuracy of detecting the irrelevant class using rotation-based SSL method vs. fully supervised models with different amounts of training labels ($2\%$, $5\%$, $10\%$, and $15\%$), with the $100\%$ baseline denoting the models trained on all available labels. Even when trained on a small amount of labeled samples, the rotation-based self-supervised model still retains high accuracy in detecting the surgically irrelevant class.}
\label{fig_f1score_nonrelevant}
\end{figure}

\begin{sidewaystable}
\tbl{The overall performance (mF1) of self-supervised learning (SSL) models fine-tuned with varied label percentages in the testing set across five-fold experiments. In contrast, fully-supervised models were trained using a ResNet-50 initialized with either ImageNet weights or random weights. Baselines were trained using all available labels. The best average test accuracy, as measured by the mF1 score, is indicated in bold, across the five-fold experiments.
\label{tab:rts_eval}}
{\begin{tabular}{llccccc}
\Hline
\multirow{2}{*}{Method} & \multirow{2}{*}{Architecture}  & \multicolumn{5}{c}{Label Percentage}  \\
\cmidrule(lr){3-7}
& &  2$\%$ & 5$\%$ & 10$\%$ & 15$\%$ & 100$\%$ (baseline)\\
\hline
SSL-rotation & ResNet-50  & \bf{96.00} ($\pm$2.09) & \bf{97.27} ($\pm$0.08) & \bf{97.48} ($\pm$1.13) & \bf{97.74} ($\pm$1.49) & \bf{98.02} ($\pm$1.00) \\
SSL-jigsaw & ResNet-50  & 65.57 ($\pm$15.24) &  82.27 ($\pm$3.60) & 87.36 ($\pm$5.31) & 87.37 ($\pm$6.81) & 92.82 ($\pm$4.32) \\
SSL-simclr & ResNet-50  & 77.72 ($\pm$10.06) & 83.53 ($\pm$10.78) & 86.88 ($\pm$6.27) & 86.39 ($\pm$5.76) & 89.72 ($\pm$8.96) \\
\hline
Supervised-random init & ResNet-50 & 93.91 ($\pm$3.22) & 94.22 ($\pm$1.70) & 94.37 ($\pm$2.52) & 94.67 (($\pm$2.50) & 95.93 ($\pm$1.55) \\
Supervised-imagenet & ResNet-50 & 92.87 ($\pm$5.58) & 95.10 ($\pm$2.22) & 92.87 ($\pm$5.58) & 96.21 (($\pm$1.27) & 96.99 ($\pm$1.05) \\
\cmidrule(lr){1-7}
\hline
Color & MLP-2 & 93.54 ($\pm$4.31) & 94.54 ($\pm$4.65) &  95.24 ($\pm$4.05) & 93.72 ($\pm$4.76) & 95.56 ($\pm$2.74) \\
HOG & MLP-2 & 81.36 ($\pm$9.16) & 81.14 ($\pm$8.30) & 81.01 ($\pm$9.13) & 80.61 ($\pm$7.62) & 82.27 ($\pm$6.05)\\
Texture & MLP-2 & 60.52 ($\pm$9.95)  & 61.49 ($\pm$5.96) & 67.76 ($\pm$8.37) & 71.40 ($\pm$6.85) & 75.70 ($\pm$8.70)\\
Blob & MLP-2  & 54.92 ($\pm$8.16) & 55.40 ($\pm$7.88) & 52.59 ($\pm$6.32) & 52.66 ($\pm$6.48) & 53.06 ($\pm$7.12) \\
Fusion & MLP-2 & 79.25 ($\pm$9.84) & 79.61 ($\pm$10.32) & 81.63 ($\pm$10.73) & 85.73 ($\pm$3.38) & 88.92 ($\pm$5.08) \\
\Hline
\end{tabular}}
\end{sidewaystable}

\section{Results \& Discussion}
Our experiment aimed to evaluate the effectiveness of the proposed self-supervised learning method for detecting clinical out-of-body images in robot-assisted surgical videos. To provide a benchmark for our design choices, we also trained a fully-supervised model with a ResNet-50 backbone.
To further explore the possibility of building performant models from limited labels, we experimented with multiple sets of available labels in the training set when fine-tuning and measured their accuracy in the test set, in every fold. Table~\ref{tab:trainset_percentage_summary} shows the details of label percentage and corresponding number of frames from the training set in five folds.
Four different label percentages were considered: $2\%$, $5\%$, $10\%$, and $15\%$, which are equivalent on average to 736, 1803, 3608, and 5412 frames sampled at 1 fps across five folds, respectively. In contrast, $100\%$ denotes the baseline given full availability of training set labels in every fold.

\begin{sidewaystable*}
\tbl{The average precision and recall scores of the positive (surgically irrelevant) class for SSL methods, fully supervised methods, and basic visual representations. The top performing model in terms of precision and recall metrics of the positive class across five folds is denoted in bold.
\label{tab:rts_precison_recall_irrelevant}}
{\begin{tabular}{llllllll}
\Hline
\multirow{2}{*}{Metric} &  \multirow{2}{*}{Method} & \multicolumn{5}{l}{Label Percentage}  \\
\cmidrule(lr){3-7}
 & & $2\%$   & $5\%$     & $10\%$    & $15\%$    & $100\%$ (baseline) \\
\hline
\multirow[t]{8}{*}{Precision$_{irrelevant}$} & SSL-rotation & \bf{91.53} ($\pm$5.94) & \bf{93.15} ($\pm$2.44) & \bf{93.46} ($\pm$1.90) & \bf{94.36} ($\pm$3.70) &\bf{93.96} ($\pm$3.91)  \\

& SSL-jigsaw & 32.91 ($\pm$ 27.45) & 73.61 ($\pm$ 12.44) & 73.15 ($\pm$ 9.09) & 74.51 ($\pm$ 11.65) & 80.53 ($\pm$ 12.08) \\
& SSL-simclr & 57.10 ($\pm$ 24.88) & 71.30 ($\pm$ 5.75) & 81.66 ($\pm$ 7.11) & 72.79 ($\pm$ 9.61) & 78.08 ($\pm$ 9.77) \\
& Supervised-random init  & 85.42 ($\pm$12.10) & 87.75 ($\pm$8.62) & 84.89 ($\pm$8.35) & 86.36 ($\pm$6.23) & 88.12 ($\pm$4.58) \\
& Supervised-imagenet  & 81.70 ($\pm$15.78) & 88.71 ($\pm$5.94) & 88.37 ($\pm$5.80) & 89.37($\pm$4.26)  & 90.44 ($\pm$3.61)   \\
& Color & 88.74 ($\pm$14.94) & 86.33 ($\pm$14.16) & 87.43 ($\pm$12.63) & 82.93 ($\pm$14.89) & 87.50 ($\pm$8.62)  \\
& HOG & 78.31 ($\pm$9.51) & 75.88 ($\pm$8.9
3) & 76.50 ($\pm$7.72) & 72.51 ($\pm$9.36) & 73.80 ($\pm$9.36) \\
& Texture & 87.89 ($\pm$10.07) & 84.60 ($\pm$10.07) & 92.31 ($\pm$2.88) & 93.60 ($\pm$1.30) & 94.58 ($\pm$3.03)  \\
& Blob & 14.81 ($\pm$10.99) & 18.77 ($\pm$14.54) & 11.16 ($\pm$6.79) & 11.96 ($\pm$6.59) & 11.71 ($\pm$8.00)  \\
& Fusion & 77.27 ($\pm$10.91) & 79.82 ($\pm$8.24) & 82.77 ($\pm$4.30) & 79.27 ($\pm$7.87) & 84.02 ($\pm$10.90) \\
\hline
\multirow[t]{8}{*}{Recall$_{irrelevant}$} & Self-supervised (Rot. pretrain) & \bf{93.76} ($\pm$5.68) & \bf{96.77} ($\pm$3.02) &  \bf{97.03} ($\pm$3.19) & \bf{97.20} ($\pm$2.37) & \bf{98.81} ($\pm$0.50)    \\
& SSL-jigsaw & 36.02 ($\pm$ 31.26) & 66.70 ($\pm$ 18.13) & 85.05 ($\pm$ 18.21) & 84.29 ($\pm$ 19.73) & 95.12 ($\pm$ 2.62) \\
& SSL-simclr & 60.97 ($\pm$ 14.18) & 73.16 ($\pm$ 27.15) & 74.40 ($\pm$ 21.55) & 82.03 ($\pm$ 19.13) & 87.29 ($\pm$ 15.61) \\
& Supervised-random init   & 93.39 ($\pm$5.51) & 91.60 ($\pm$5.94) & 94.97 ($\pm$3.94) & 94.05 ($\pm$5.25) & 97.13 ($\pm$2.63) \\
& Supervised-imagenet &  93.56 ($\pm$6.34) & 93.41 ($\pm$7.33) & 94.40 ($\pm$7.20) & 96.84 ($\pm$2.79) & 98.65 ($\pm$0.66)  \\
& Color  & 89.29 ($\pm$8.97) & 95.11 ($\pm$4.60) &96.19 ($\pm$2.60) & 96.05 ($\pm$2.53) & 96.71 ($\pm$1.64)  \\
& HOG & 58.43 ($\pm$19.42) & 59.79 ($\pm$19.62) & 59.17 ($\pm$20.77) & 61.07 ($\pm$20.02) & 66.13 ($\pm$19.03) \\
& Texture & 15.06 ($\pm$14.46) & 15.73 ($\pm$8.12) & 25.07 ($\pm$12.95) & 29.76 ($\pm$11.33) & 39.14 ($\pm$14.34)  \\
& Blob & 17.12 ($\pm$18.24) & 28.88 ($\pm$27.56) & 24.50 ($\pm$31.86) &  24.24 ($\pm$32.14) & 24.01 ($\pm$31.62) \\
& Fusion & 54.42 ($\pm$21.34)  & 53.80 ($\pm$22.61) &  58.05 ($\pm$23.26)  & 67.16 ($\pm$5.81) & 74.52 ($\pm$7.83) \\
\Hline
\end{tabular}}
\end{sidewaystable*}

\subsection{Comparison of self-supervised algorithms}
In recent years, a plethora of self-supervision algorithms have been proposed to tackle the challenge of feature extraction without the use of labels. While it is impractical to compare the performance of every algorithm, we sought to investigate the efficacy of two highly successful self-supervision algorithms: Jigsaw~\cite{noroozi2016unsupervised}, which involves shuffling and unshuffling pieces of an image as an auxiliary task, and SimCLR~\cite{chen2020simple}, which utilizes a contrastive learning approach. We pre-trained models using these algorithms on the same datasets that were used for rotation pre-training, and subsequently fine-tuned the models using the same algorithm.

The rotation-based self-supervised method outperformed the Jigsaw and SimCLR algorithms when fine-tuning with only $2\%$ of labels. Specifically, the rotation-based SSL method achieved an mF1 score of $96\pm2.09$, while Jigsaw achieved only $65.57 \pm 15.24$, and SimCLR slightly better with $77.72 \pm 10.06$. Both Jigsaw and SimCLR had lower average scores and were more inconsistent than the rotation-based SSL method.As more labels were added, Jigsaw and SimCLR improved, reaching mF1 scores of $82.27 \pm 3.60$ and $83.53 \pm 10.78$, respecically. However, they still fell significantly short of the performance achieved by the rotation-based method. While Jigsaw and SimCLR continue to improve with additional labels used for fine-tuning, and can match some of the image feature-based algorithms, they are never able to match the performance of the rotation-based SSL method. These results demonstrate that the rotation-based self-supervision method is a strong choice for this particular framework of self-supervision use case.

Although the proposed rotation-based self-supervision method was designed to simulate the movement and variability present in endoscopic images, it is important to acknowledge that there may be other factors contributing to its superior performance compared to other successful self-supervision methods in different domains. One potential explanation is the relatively small size of our pretraining dataset, particularly when compared to standard self-supervision benchmarks like ImageNet, which could result in Jigsaw and SimCLR requiring more data to achieve peak performance. Nonetheless, a thorough investigation of the reasons behind the observed performance differences between the rotation-based method and other self-supervision algorithms is beyond the scope of this study.

\subsection{Comparison of fully-supervised algorithms}
Table~\ref{tab:rts_eval} shows the comparison of our proposed method with a set of labeled data against fully-supervised methods for detecting irrelevant segments in surgical videos from the test set. 
Two fully-supervised approaches were considered: (1) taking the model pre-trained on ImageNet as an initialization (ImageNet pretrain), and (2) training the model from scratch without any pre-training (Random Init.).

When using all labels from the training set, we found that self-supervised models achieved the best overall performance with the highest average mF1 score of $98.02 \pm1.00$, compared to fully-supervised models and other previous methods. Using only $5\%$ of labels in training set, the self-supervised method retains a high performance with average mF1 score of $97.27 \pm0.08$, which is as performant as both fully-supervised models using all the labels in training set. For the fully-supervised approach using the same evaluation setup, training the model with random initialization (mF1 $=95.93 \pm1.55$) performs nearly as well as the one initialized with pre-trained ImageNet weights (mF1 $=96.99 \pm1.05$), if not better. 
The results indicate that self-supervised learning can effectively learn representations of endoscopic images from a large amount of unlabeled data. Such an approach is beneficial especially when limited labels are available for fine-tuning.

As clinically irrelevant images pose a high risk of exposing sensitive information in particular when the endoscope is pulled out of a patient's body, accurately classifying these irrelevant images from a large amount of regular inside body views is critical for success. Here, we specifically focus on the performance of the positive class and report its recall and precision values accordingly. Fig.~\ref{fig_f1score_nonrelevant} shows the average F1 score of the irrelevant class across five folds. Table~\ref{tab:rts_precison_recall_irrelevant} compares the average precision and recall scores of the irrelevant class from the self-supervised model and the fully-supervised models. Fig.~\ref{fig_f1score_nonrelevant} shows that the self-supervised model consistently provides the highest average F1 scores of irrelevant images, given varied amounts of available labels. 
More specifically, as shown in Table~\ref{tab:rts_precison_recall_irrelevant}, fully-supervised models only achieve precision ranging from $81.70 \pm15.78$ to $85.42 \pm 12.10$ on average for the surgically irrelevant class, when trained on $2\%$ of the labels. In contrast, the precision of the self-supervised models reaches $91.53 \pm5.94$ for detecting the irrelevant class when trained on $2\%$ of the labels, and smaller variances in precision are observed across five folds. This result shows that the self-supervised model can not only provide a high recall for irrelevant frames, but also performs significantly better in generating true positive predictions among all other models. This could be a difficult task due to the large complexity of surgical scenes, patient anatomy, and the presence of visual challenges in surgical videos (we discuss further in Section~\ref{sec:error_analysis}).

\begin{figure}[tb]
\centering
\includegraphics[width=0.5\linewidth,clip,trim=18pt 10pt 50pt 0pt]{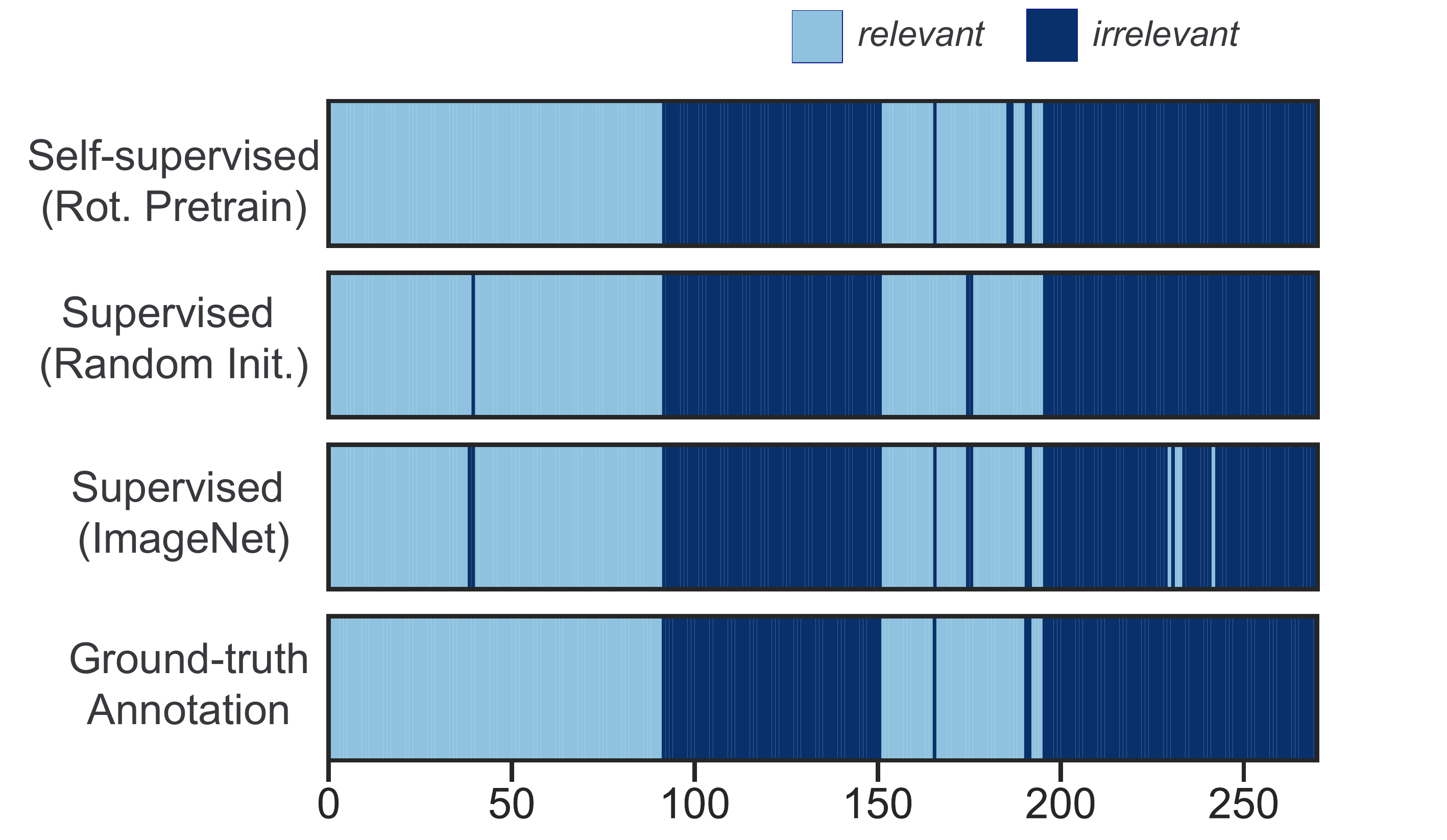}
\caption{Qualitative visualization of sequential prediction performance in a sample video. The x axis is the timestamp of individual frames in seconds. The predicted class labels of the video frames are color-coded as light blue for relevant frames and dark blue for irrelevant frames. The color bar corresponds to the prediction output from each model that was trained using $5\%$ labels and the ground-truth annotations for comparison.}
\label{fig_video_level_label_comp}
\end{figure}

While our main focus of this work is to develop a frame-based classification approach, we did investigate how the resulting models perform sequentially at the video clip level and whether or not further extracting additional temporal information from a set of frames is needed for modeling. In Fig.~\ref{fig_video_level_label_comp}, we present the visualization of sequential predictions obtained by applying the model at every second on an example video. For comparison, we use individual models trained with $5\%$ labels and compare the results with the ground-truth annotations. As shown, while there exist a few false positive predictions (recognizing relevant images falsely as the out-of-body class), our self-supervised model provides the most consistent performance with continuous predictions from frame to frame. In contrast, fully supervised models have less smooth, noisy predictions at the video level, in particular when misclassifying the irrelevant class as relevant.

\begin{figure*}[tb]
\centering
\includegraphics[width=0.9\linewidth,clip,trim=5pt 0pt 5pt 0pt]{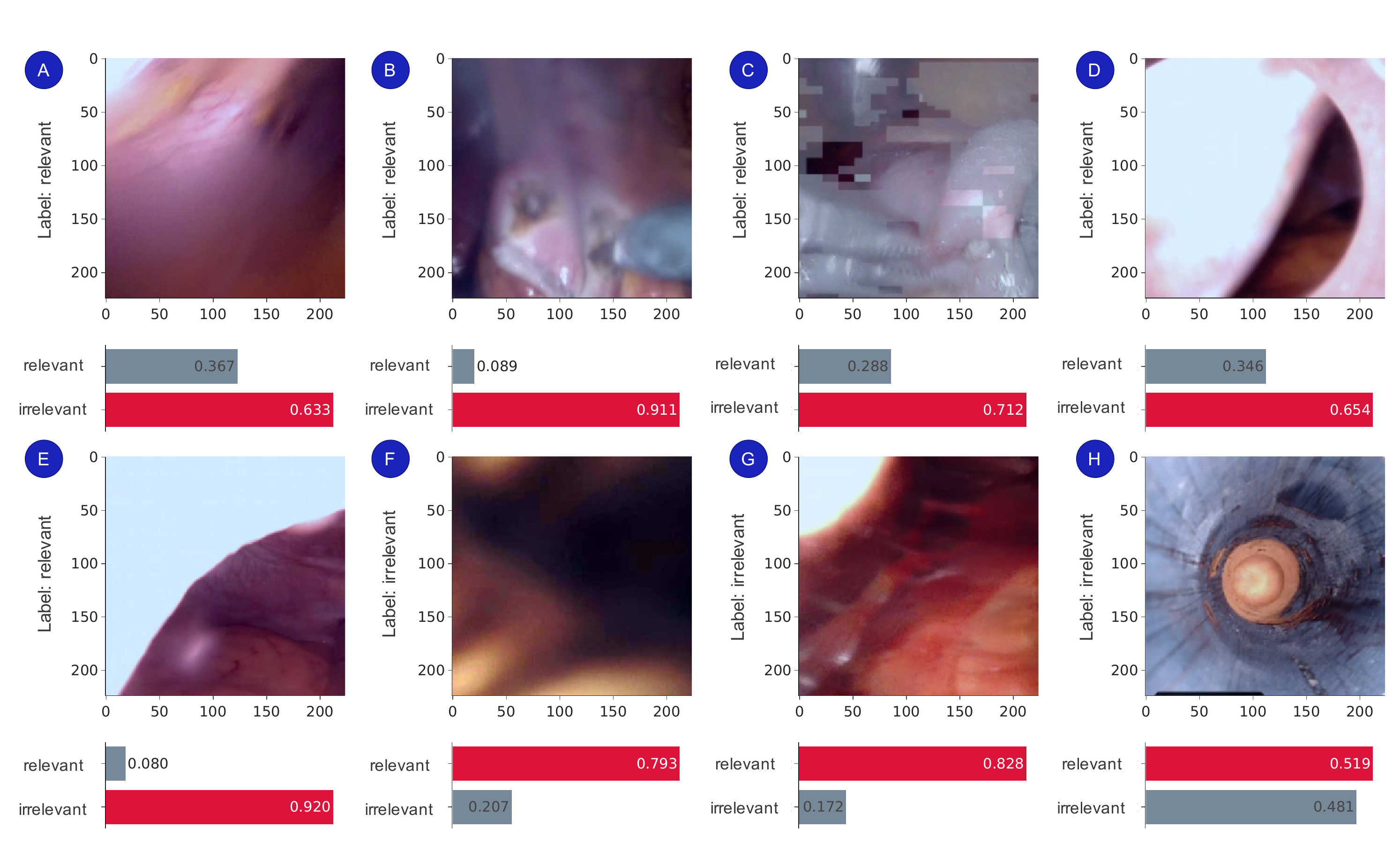}
\caption{Examples of endoscopic video frames comparing self-supervised predictions vs. ground truth annotations: (A) missed relevant frame with endoscope camera stains, (B) missed relevant frame with heavy smoke,  (C) missed corrupted frame with signal noise, (D, E, H) missed inside-cannula frames with partial in-body views, and (F, G) relevant frames mislabeled as irrelevant. The red colored bar denotes the falsely predicted class label. The number shows the confidence of prediction for each class. Several visual challenges were encountered in endoscopic imaging, such as blur, heavy smoke, signal noise, out-of-focus, etc. }
\label{fig_error_analysis}
\end{figure*}

\subsection{Comparison of visual representations}
While conventional visual representations, such as color histogram, Histogram of Oriented Gradient (HOG), Texture, etc., have been used in a wide range of computer vision applications~\cite{munzer2018content,stanek2012automatic,ghosh2018chobs,akgul2011content,oh2007informative}, their role for detecting irrelevant out-of-body segments in a surgical context is unclear.
For comparison to our framework, we explored the performance of four basic image features (Color, HOG Contour, Texture, and Blob) individually as a means of testing whether handcrafted feature extraction could obtain comparable performance. We also obtained a combination of the four basic representations (Fusion) by concatenating them into a 1-D feature vector as the input to train a classification model.
As shown in Table~\ref{tab:rts_eval}, the performance of models using basic handcrafted representations, while being comparable, was lower than both representations learned by self-supervised and fully-supervised models. The color representation feature in the image achieved the best accuracy ($95.56 \pm2.74$ mF1), followed by HOG-based contour feature ($82.27 \pm6.05$ mF1) and Texture feature with ($75.70 \pm8.70$ mF1).  
This suggests that, despite suffering from a lack of explainability compared to basic image descriptors, the self-supervised approach benefits from the representations learned during the pre-training stage, even though no labels were used, and thus could offer substantially better generalizability to various endoscopic views in unseen videos, compared to handcrafted visual representations.

In addition, we observed that models trained on basic visual representations tend to have a worse performance when less labels are available for training. Table~\ref{tab:rts_eval} shows the overall accuracy of the color-based model dropped to an average $93.54 \pm 3.22$ mF1 when training on $2\%$ labels. However, as shown in Table~\ref{tab:rts_precison_recall_irrelevant}, while the recall score was improved from $89.29\%$ to $96.19\%$, the precision for detecting the irrelevant images remained relatively low (ranging $82.93\%$ to $88.74\%$), given an increase of data labels from $2\%$ to $15\%$. It indicates the color-based features might be challenging to generalize to endoscopic images that have very different characteristics due to heavy smoke, high contrast, partial in-body views, etc.
The texture-based model was not able to maintain high performance ($60.52 \pm9.95$ mF1) with $2\%$ of labels, but it achieved an average $17.98\%$ performance boost and reached $71.40\pm6.85$ mF1 when labels increased to $15\%$. In general, the larger amount of labels improved the performance of the image feature-based models by a large margin, however, it could become more expensive and less practical to train them, considering a significant amount of effort and cost would be required for labeling. 

It is worthwhile to note that, compared to individual features, the visual fusion model which combined basic representations (Color+HOG+Texture+Blob) simply captured more visual information but did not necessarily lead to a better performance. While the visual fusion was able to improve the overall performance with average mF1s ranging from $79.25\pm9.84$ to $85.73\pm3.38$ when training on $2\%$ to $15\%$ labels (baseline $88.92 \pm5.08$ mF1), the fusion method provided less accurate results especially when identifying the irrelevant class, compared to the color-based models. Several visual representations such as Texture and Blob features can more likely overfit the training data (Texture with high precision and low recall, Blob with low precision and low recall for the positive class), and thus the fusion model was not able to improve much and generalize well to a large amount of endoscopic images. 

\subsection{Error Analysis}
\label{sec:error_analysis}
In addition to quantitative performance evaluation, we investigated the classification error and discussed the potential visual challenges that we have found in typical surgical video recordings. Fig~\ref{fig_error_analysis} shows examples when the proposed model misclassified the frames with a high confidence score. It can be observed that classification accuracy can be affected by the ambiguity of the transition between irrelevant and relevant segments, label consistency and various visual challenges.

In particular, some segments might not have a clear visual definition when the endoscope quickly passes through the cannula, and these video segments can be classified as either in-body (relevant) or out-of-body (irrelevant). For instance, Fig.~\ref{fig_error_analysis} (E) shows a confusing relevant image with a relatively large in-body view. On the contrary, Fig.~\ref{fig_error_analysis} (D) only has a limited view of inside body and can be viewed as an irrelevant out of body frame. 
We also observed that several discrepancies existed between our annotation labels and the ground truth. Fig.~\ref{fig_error_analysis} (G) is falsely labeled due to the error in annotations, however the proposed model is able to correctly predict the truth with a high confidence of 0.828 as the surgically relevant class. This might be because of the very short duration of the segment, which is likely ignored when annotating the full-length video. In addition, we observe several visual challenges, such as endoscope stains, motion blur, smoke, and signal noise as shown in Fig.~\ref{fig_error_analysis} (A), (B), and (C), that are most likely present in inside-body, surgically relevant samples. It should be noted that the majority of misclassified frames are false positives, i.e., the true inside-body relevant samples are misclassified as the irrelevant class. As shown in Table ~\ref{tab:rts_precison_recall_irrelevant}, the precision score of the irrelevant class when given $100\%$ labels with self-supervised pre-training is slightly less than when given $15\%$ labels.
This can be explained by that fact that the random presence of visual challenges could make it difficult to infer the correct class labels, especially for these surgically relevant samples that visually might look very similar to the irrelevant ones. 
However, our proposed self-supervised model is able to detect the positive out-of-body class with the highest precision and recall scores, which reduces the effect of noisy predictions caused by the visual challenges while protecting sensitive clinical information often shown in out-of-body video segments.

\subsection{Limitations}
In this study, we performed experiments exclusively on endoscopic videos in robot-assisted surgery collected from da Vinci X and Xi surgical systems. Due to limited case volumes of surgical procedures in our dataset, we did not compare performance across multiple procedures or specialities. For future work, it could be useful to further explore the generalizability of the proposed method on recordings from different types of procedures or surgeries, such as laparoscopic surgery. Furthermore, we adopted a rotation-based SSL method for model pretraining in this paper, and compared it to two other methods: Jigsaw and SimCLR. In the future, other transformations such as color permutation \cite{jing2020self}, or other contrastive learning approaches \cite{chen2020big} could be considered to further validate the choice of a rotation-based method. While conceptually simple, the rotation-based pretraining shows its effectiveness in learning visual representations without any supervised labels. Especially for videos in the surgical field, we believe applying random rotation transformation to endoscopic images effectively simulates the stochastic nature of endoscope movement and thus results in a better performance when fine-tuning for the final detection. In addition, it would be interesting to further investigate the effects of self-supervised methods when applied to different network architectures, such as vision transformer~\cite{dosovitskiy2020image}, for learning visual features. Lastly, despite the competitive results of our frame-wise detection, self-supervised video-based feature learning based on a sequential set of frames could be considered to explicitly capture temporal information contained in surgical videos. Nevertheless, we hope our study can inspire researchers to consider using self-supervised methods to learn endoscope-specific characteristics for specific surgical applications.

\section{Conclusion}
As minimally-invasive robot-assisted surgery becomes increasingly prevalent, it is crucial to ensure the safe and responsible handling of large-scale surgical video recordings. This is essential to enable cross-institutional data sharing, educational applications, and collaborations among surgical communities. However, manually reviewing and annotating full-length surgical videos is often prohibitively time-consuming and challenging. To address this issue, automated methods that can efficiently clean sensitive video frames and remove surgically irrelevant information are needed.

In this study, we present a high-performing framework that leverages self-supervised learning with minimal labels to automatically detect out-of-body segments in surgical videos. We validate the framework on a clinical video dataset collected from robot-assisted surgeries. The framework features self-supervised pre-training without using any labels, followed by supervised fine-tuning with a limited set of labels. The resulting model achieved highly accurate and robust performance, requiring only $2\%$ labeled data, significantly reducing its dependency on labeling while maintaining high-level classification performance.

Our results highlight the potential of automatic privacy protection as a foundational component of surgical video management and processing in clinical and technical communities. By automating the cleaning of sensitive video frames and removing surgically irrelevant information, our framework paves the way for more efficient surgical video data management and subsequent analysis.

\section*{Acknowledgment}
The authors would like to thank Kiran Bhattacharyya, Aneeq Zia, Samuel Bretz, Yachna Sharma, and Linlin Zhou for thoughtful discussions and support around methodologies. The authors would also like to thank Imogen Boaler Farlie for her data annotation support.

\bibliographystyle{ws-jmrr}

\end{document}